\documentclass[journal,comsoc]{IEEEtran}
\usepackage[T1]{fontenc}
\usepackage{cite}

\ifCLASSINFOpdf
   \usepackage[pdftex]{graphicx}
\else
   \usepackage[dvips]{graphicx}
\fi

\usepackage{amsmath}
\usepackage[cmintegrals]{newtxmath}

\usepackage[numbers]{natbib}
\usepackage{booktabs}
\usepackage{multirow}
\usepackage{multicol}
\usepackage{amsmath}
\usepackage{algpseudocode}
\usepackage{mathtools}
\usepackage[linesnumbered,ruled,vlined]{algorithm2e}

\mathchardef\mhyphen="2D

\begin{document}
%
\title{Continual Growth of Autonomous Robots through Independent and Parallel Large-Small Model Collaboration}

\author{Hong~Su
\IEEEcompsocitemizethanks{\IEEEcompsocthanksitem H. Su is with the School of Computer Science, Chengdu University of Information Technology, Chengdu, China.\\
 E-mail: suguest@126.com. \\
\protect\\
}
\thanks{}}

\markboth{Journal of \LaTeX\ Class Files,~Vol.~14, No.~8, August~2015}%
{Shell \MakeLowercase{\textit{et al.}}: Bare Demo of IEEEtran.cls for IEEE Communications Society Journals}
%

\maketitle

\begin{abstract}
Continual growth through environmental interaction is essential for autonomous robots to expand their knowledge and capabilities beyond their initial design. However, existing approaches often focus on predefined tasks or rely on large language models (LLMs) for reasoning without systematically converting thinking results into reusable interaction capabilities. This paper proposes an independent and parallel large-small model collaboration framework for continual autonomous growth. Small models continuously interact with the environment, while the large model autonomously selects thinking subjects and reasons about current experiences, historical interactions, and previous thinking processes. Through such reasoning, the robot discovers new object properties, interaction methods, and learning opportunities. The resulting hypotheses guide evidence collection, training, and validation, leading to the formation and improvement of multiple small models. These models enhance subsequent environmental interaction, thereby establishing a continual growth cycle of interaction, thinking, and learning. Previously acquired models are retained for future reuse. Simulation results show that the proposed framework improves frozen-model accuracy from 58.10\% to 70.14\% in changing environments and expands mastered object families from one to two, although non-LLM adaptive exploration masters three.
\end{abstract}


\IEEEpeerreviewmaketitle

\section{Introduction}
\label{sec:introduction}

Autonomous robots operating in changing and unfamiliar environments need not only to perform existing tasks reliably but also to continuously expand their knowledge and interaction capabilities. During routine activities, a robot may encounter new objects, observe unfamiliar phenomena, or discover that an alternative interaction reveals information unavailable through its current sensing methods. Interactive perception and autonomous exploration demonstrate that such activities can provide valuable learning experiences \cite{bohg2017interactive,forestier2022imgep}. Large language models (LLMs) offer flexible reasoning capabilities for interpreting these experiences, while lightweight learned models enable efficient and repeated execution. Combining these capabilities offers an opportunity for robots to grow beyond their initially provided knowledge through continued environmental interaction.

However, existing approaches do not fully address how independently conducted reasoning can be converted into an expanding repertoire of learned interaction capabilities during routine robot operation. Hierarchical and dual-system architectures combine high-level reasoning with specialized action models and can separate their execution frequencies \cite{huang2025thinkact,dey2026arp}, but their primary emphasis is on planning and executing tasks. Other approaches use LLMs to discover environmental rules or facilitate policy learning \cite{zhu2025psalm,zhou2025walle}, while continual-learning and skill-library approaches support knowledge retention and the accumulation of reusable skills \cite{wang2024voyager,tziafas2024lrll}. Despite these advances, the integration of autonomous thinking-subject selection, interaction-based evidence collection, and the continual formation of lightweight models remains an important problem. In particular, a useful reasoning result does not necessarily become a capability that the robot can subsequently exercise without further LLM inference.

In this paper, we propose an independent and parallel large--small model collaboration framework for the continual growth of autonomous robots. The Small Model, representing a dynamic collection of lightweight models, performs routine tasks, interacts with the environment, and learns from feedback. The Big Model independently selects thinking subjects from current observations, the activities of the Small Model, historical interaction experiences, and its previous thinking processes. It can investigate ongoing activities or issues that the Small Model has not addressed, without requiring the latter to wait for every reasoning result. Its suggestions are treated as candidate hypotheses or methods rather than mandatory action commands.

Through continued interaction, the Big Model can identify new object properties, informative interaction methods, and opportunities for learning. The robot then collects empirical evidence, refines candidate methods, and trains and validates small models that capture the discovered relationships. Existing models can be extended, and additional models can be formed when new capabilities are needed. Previously acquired models are retained for reuse or reactivation when applicable. The growing model repertoire thereby provides resources for subsequent environmental interaction, which in turn generates further experiences for thinking and learning. The objective is to establish a continual growth cycle rather than merely improve the accuracy of a fixed task model.

The main contributions of this paper are as follows:
\begin{enumerate}
    \item An independent and parallel large--small model collaboration mechanism that enables routine interaction and autonomous thinking to proceed separately, with bidirectional information exchange and nonmandatory reasoning proposals.
    \item A thinking-driven model formation mechanism that converts reasoning about environmental interactions into empirically validated small models, supporting both the selection of informative interaction methods and the acquisition of new task capabilities.
    \item A continual growth framework that integrates the accumulation, retention, and context-dependent reuse of small models with subsequent environmental interaction. Simulation experiments demonstrate the feasibility of knowledge acquisition and reactivation, and show an increase in mastered object families from one to two compared with fixed initial sensing, while a non-LLM adaptive exploration baseline masters three.
\end{enumerate}

The remainder of this paper is organized as follows. Section~\ref{sec_related_work} reviews related work. Section~\ref{sec:model} presents the proposed independent and parallel thinking and learning model. Section~\ref{sec:verification} describes the experimental settings and evaluates adaptation, knowledge reuse, and cumulative model growth. Finally, Section~\ref{sec_conclusion} concludes the paper.

\section{Related Work}
\label{sec_related_work}

\subsection{Large--Small Model Collaboration and Autonomous Reasoning}

Large language models (LLMs) offer semantic knowledge and reasoning capabilities that complement the specialized perception and control models used in robotics. SayCan combines language-based skill selection with learned affordance estimates to ground high-level instructions in the robot's executable capabilities \cite{ichter2023saycan}. Hi Robot uses hierarchical vision--language--action models to interpret complex instructions and feedback and execute the resulting subtasks \cite{shi2025hirobot}. These approaches benefit from the complementary strengths of semantic reasoning and specialized control, while reducing the risk of selecting physically infeasible actions. However, their primary objective is to accomplish a given task using available capabilities. The continual acquisition of additional small models from independently selected thinking subjects is not the main focus of these architectures.

Recent dual-system architectures further separate deliberation from action execution. ThinkAct compresses reasoning plans into visual latent representations that condition a downstream action model, facilitating long-horizon planning and adaptation \cite{huang2025thinkact}. More recently, Asynchronous Robot Policies (ARP) decouples semantic reasoning from high-frequency action generation and explicitly addresses the temporal mismatch between updated observations and older semantic information \cite{dey2026arp}. Such mechanisms improve reactivity and provide a practical alternative to executing an expensive reasoning model within every control step. Nevertheless, the separation of inference frequencies does not by itself provide a mechanism for deciding what new environmental characteristics deserve investigation or for forming persistent task models from the results of such investigations.

Experience-based reflection provides another way to improve LLM agents. Reflexion stores feedback-derived verbal reflections in episodic memory to improve later attempts without updating the LLM parameters \cite{shinn2023reflexion}. Its advantage is that previous failures can influence subsequent decisions without expensive fine-tuning. However, the improvement primarily depends on the agent's use of reflective memory during later reasoning, rather than on the creation of lightweight models that can independently exploit the acquired knowledge. In the proposed framework, earlier thinking is itself a source of experience, but its outcomes may also identify new learning objectives for the Small Model.

Unlike a task-level planning hierarchy, the proposed framework treats thinking and routine interaction as independently scheduled activities. The Big Model may investigate ongoing activities, revisit earlier events, or select issues that the Small Model has not addressed. Its suggestions are candidates rather than mandatory action commands. The distinguishing objective is therefore not merely faster execution through parallel inference, but the conversion of autonomous thinking into lasting interaction capabilities while routine activities continue. The present implementation realizes concurrency within scheduling blocks; it does not claim the finer-grained asynchronous control demonstrated by ARP.

\subsection{Interaction-Based Discovery and Thinking-Guided Learning}

Interactive perception exploits the coupling between physical actions and sensory observations: changing the viewpoint or manipulating an object can reveal information that passive perception cannot obtain \cite{bohg2017interactive}. Intrinsically Motivated Goal Exploration Processes (IMGEP) further enable autonomous agents to generate learning goals, select them using intrinsic rewards, and reuse experience across goals \cite{forestier2022imgep}. These approaches demonstrate that active exploration can create informative experiences and support increasingly complex skills without relying on an LLM. Their effectiveness, however, depends on the available representations of goals, actions, and exploration incentives. The proposed framework investigates a complementary route in which semantic reasoning can suggest which previously unexamined interaction or property should be investigated and what evidence would support the hypothesis.

LLMs have also been employed to acquire explicit knowledge of environmental dynamics. PSALM combines LLM-generated hypotheses, executed plans, and observed outcomes to infer logical action semantics for symbolic planning \cite{zhu2025psalm}. WALL-E extracts symbolic knowledge from exploration trajectories and encodes it into executable representations that support world-model-based LLM planning \cite{zhou2025walle}. Both approaches demonstrate the benefit of grounding reasoning in environmental evidence rather than relying exclusively on the prior knowledge of an LLM. Their principal knowledge representations are symbolic action descriptions or executable world-model knowledge, whereas the present framework emphasizes training lightweight predictive models from measured observations and delayed feedback. These representations are complementary: symbolic knowledge is useful for interpretable planning, while learned predictors can capture empirical relationships without requiring every relationship to be expressed as an explicit rule.

LLMs can also contribute to the training process itself. Eureka uses iterative language-model generation and evaluation of reward code to train reinforcement learning policies \cite{ma2024eureka}. AutoRT uses foundation models to propose context-dependent robotic tasks and coordinate data collection, demonstrating the value of diverse interaction data for subsequent policy learning \cite{ahn2024autort}. These studies show that the benefit of an LLM can extend beyond immediate action selection to the acquisition of training experience and the improvement of learned policies. However, reward generation for specified learning problems and large-scale data collection do not necessarily address the same problem as the continuous, local formation and reuse of specialized models during a robot's routine operation.

The proposed approach connects interaction-based discovery with model formation. The Big Model can hypothesize that an alternative interaction, such as changing illumination, rotating an object, or measuring a response after contact, reveals useful information. The Small Model can revise the candidate method, learn its empirical usefulness, and collect independently verified training evidence before adopting a new task capability. This differs from treating LLM-generated predictions or rules as sufficient evidence of correctness. An important limitation is that useful hypotheses still require adequate measurement opportunities: semantic reasoning alone cannot guarantee learning, and non-LLM adaptive exploration remains a meaningful alternative in a finite, structured interaction space.

\subsection{Continual Learning and the Growth of Reusable Capabilities}

Continual learning addresses the acquisition and retention of knowledge from sequential experiences. Lesort et al. review its application to robotics, emphasizing the difficulties arising from changing data distributions, knowledge retention, and continuous interaction \cite{lesort2020continual}. LIBERO provides a lifelong robot learning benchmark for evaluating the transfer of declarative and procedural knowledge across manipulation tasks \cite{liu2023libero}. These studies offer important methods and evaluation principles for avoiding interference between learned capabilities and assessing transfer. Nevertheless, learning across a supplied sequence of tasks is distinct from deciding, during interaction, which new capabilities should be learned and what observations are required to acquire them. The latter question motivates the integration of autonomous reasoning and continual learning in this work.

Open-ended embodied agents also accumulate reusable skills. Voyager combines an automatic exploration curriculum, a growing library of executable skills, and iterative improvement from environmental feedback in Minecraft \cite{wang2024voyager}. Lifelong Robot Library Learning (LRLL) similarly uses self-guided task exploration, retrieval of past experience, and skill abstraction to expand a library of composable robotic behaviors \cite{tziafas2024lrll}. These approaches are particularly relevant because previously acquired skills provide resources for acquiring more complex skills. Their advantages include compositionality and the preservation of reusable behaviors without requiring continual gradient-based optimization of the LLM. Their emphasis, however, is on executable behavioral skills and code libraries constructed through LLM reasoning, rather than the continual training of measurement-selection and task-prediction models from new physical observations.

The proposed framework instead considers the Small Model as a dynamic repertoire of lightweight models. New capabilities can be formed when previously available observations or predictors are insufficient, while existing capabilities can be improved or reused when they remain applicable. A learned capability that becomes temporarily unsuitable is retained and can be reactivated after fresh validation. This distinguishes the retention of model parameters from the decision to apply those parameters in a particular situation. It also connects continual learning to the acquisition of interaction knowledge: the robot learns not only the characteristics of objects, but also which interactions reveal those characteristics and how the resulting observations can be interpreted independently of the LLM.

Taken together, the reviewed approaches establish the importance of reasoning--action separation, autonomous exploration, feedback-based learning, and reusable skill accumulation. The proposed framework combines these directions around a specific growth cycle: continued interaction provides new experiences, independently scheduled reasoning identifies potential learning opportunities, and empirical learning forms small models that support subsequent interaction. Its intended distinction is the integration of these mechanisms rather than the novelty of any one component in isolation. The current experiments examine the formation, independent use, accumulation, and reactivation of learned capabilities; whether those capabilities themselves lead to the discovery of previously inaccessible objects remains a question for future verification.

\section{Independent and Parallel Thinking and Learning Model}
\label{sec:model}

The proposed model aims to enable an autonomous robot to grow through continued interaction with its environment, rather than merely to improve performance on a fixed task. It employs two independently operating components: a \emph{Small Model}, responsible for routine task processing, environmental interaction, and the learning of reusable capabilities, and a \emph{Big Model}, responsible for flexible reasoning about current and historical experiences. The Small Model denotes a collection of lightweight models rather than a single fixed predictor, whereas the Big Model is implemented using a large language model (LLM). Initially, the collection may contain only the capabilities needed for basic interaction. As the robot encounters additional objects, phenomena, and consequences of its actions, the Big Model can identify opportunities for learning and facilitate the formation or extension of small models that capture the characteristics of these interactions. The objective is to gradually enrich the robot's repertoire of independently usable capabilities, not merely to increase the accuracy of one classifier. Their relationship is not a conventional hierarchy in which the Big Model directs every action of the Small Model. Instead, each component has its own activities and can use the processes and results of the other to improve subsequent activities. In particular, the Big Model can discover new methods of obtaining environmental information, while the Small Model can further improve these methods and convert their verified results into independently usable task knowledge. Continued interaction then provides new experiences for both components.

\subsection{Independent and Parallel Operation}
\label{sec:model_parallel}

The Small Model operates continuously using its existing capabilities. Its activities include perceiving the environment, selecting and executing actions, making task decisions, and learning from the resulting feedback. The Big Model operates on a separate reasoning path and is not restricted to answering requests issued by the Small Model. It can examine an activity that the Small Model is currently performing, reconsider a previous activity, or investigate an issue that the Small Model has not addressed. Thus, an apparently successful action may also become a subject of thinking, for example, when its efficiency could be improved or when its success depends on an insufficiently understood environmental condition.

The input to the Small Model consists of environmental observations, the internal state of the robot, and available information from the Big Model, such as newly received hypotheses, its current reasoning status when available, and suggestions for learning or interaction. The Big Model receives environmental observations, the robot's internal state, and information about the Small Model's activities, including actions, predictions, learning processes, and feedback. These inputs may also describe the severity of an observed problem, such as potential damage, repeated errors, or unfavorable consequences. Severity is evidence for deciding what deserves attention, rather than a mandatory rule that triggers Big Model intervention.

Information exchange does not imply compulsory execution. The Small Model can continue its routine activities while the Big Model is thinking, and can accept, defer, revise, or leave unused a received proposal according to its current knowledge, evidence, and available resources. Likewise, the Big Model may choose not to investigate a particular Small Model activity. The two components share the robot's physical resources; executable actions must satisfy the available interfaces and resource constraints. Such coordination does not require a central reasoning component that selects every action and thinking subject.

The experimental implementation realizes this independence using concurrent reasoning and interaction within scheduling blocks. The Small Model acts using previously available knowledge while a new reasoning request is pending; the resulting proposal becomes available at a block boundary. This is block-level concurrency rather than a guarantee of completely wait-free operation. The essential property is that routine decisions do not require a fresh Big Model inference for each action.

\subsection{Autonomous and Experience-Guided Thinking}
\label{sec:model_thinking}

The Big Model should be able to determine not only \emph{how} to reason about an issue, but also \emph{what} to think about. A failure, an unfamiliar observation, an unexpected consequence of an action, or a possible improvement to an otherwise successful activity may provide a new thinking subject. Multiple subjects may coexist and may require evidence collected at different times. Therefore, the model maintains pending thinking subjects together with the currently investigated subject. A subject records its question, relevant observations, previous reasoning, evidence requirements, and current status. At a reasoning opportunity, the Big Model can continue the current subject, select another subject, suspend it until further evidence becomes available, revisit an earlier subject, or ignore an issue that does not warrant further investigation. The selection is based on the available evidence and reasoning history, not solely on a predefined error threshold or a fixed ordering of subjects.

A thinking subject need not be resolved in one reasoning cycle. For example, after observing that a color-based decision becomes unreliable, the Big Model may hypothesize that a different viewing distance or illumination reveals a more informative property. It may suggest an appropriate measurement, wait for the robot to obtain evidence and for the corresponding feedback to arrive, and subsequently revise its hypothesis. Observations obtained from different interactions can thus contribute to the same thinking process even when they are separated in time.

An interaction experience is represented as
\begin{equation}
    e_i=(s_i,p_i,x_i,\hat{y}_i,y_i,t_i),
    \label{eq:model_experience}
\end{equation}
where $s_i$ is the observable situation (e.g., location, object identity, or recent task performance), $p_i$ is the executed interaction or measurement method, $x_i$ is the resulting observation, $\hat{y}_i$ is the task prediction, $y_i$ is the subsequently available task feedback, and $t_i$ is the interaction time. An experience without released feedback may support a hypothesis, but cannot be treated as a labeled training or validation sample. The situation does not contain the unknown true environmental relationship that the robot is attempting to discover.

The Big Model also retains its own thinking experience, including why a subject was selected, which hypotheses and methods were proposed, what evidence was requested, and how later interaction supported or contradicted the proposals. Relevant records can be retrieved from the full experience archive according to the current situation, observed effectiveness, and unresolved questions. Only a selected portion needs to be included in the current reasoning context. When available, experience from another robot can also be considered as a source of candidate methods, but its applicability must be established locally.

This feedback from earlier thinking to subsequent thinking is important: the Big Model may reconsider a previously unsuccessful hypothesis when the situation changes, identify a limitation of its earlier exploration, or suggest how a known method should be modified. The resulting thinking experience is therefore useful not only for solving the original problem but also for guiding future investigation. In the proposed framework, explicit subject management supports this autonomous reasoning process; the continual-growth implementation in Section~\ref{sec:verification} instantiates it through periodically scheduled, independently generated proposals and their historical outcomes, rather than implementing every possible subject-management operation.

\subsection{Thinking-Driven Formation and Growth of Small Models}
\label{sec:model_learning}

The central mechanism is the conversion of the Big Model's thinking results into new capabilities of the Small Model. The Big Model should not repeatedly solve an encountered task on behalf of the Small Model when the corresponding capability can be acquired through learning. Instead, it can reason about why the existing capability is insufficient, what additional environmental information might be useful, how that information could be obtained, and what evidence is needed for learning. This differs from directly distilling the LLM's task predictions: the new capability is grounded in observations and feedback obtained from actual interactions.

The knowledge to be acquired is not restricted to the properties of objects. It may describe what an object is like, how an interaction reveals a property of that object, or what consequence a particular action produces in a given situation. For example, changing illumination may reveal a surface property, rotating an object may expose a previously unobservable feature, and contact followed by waiting may yield a material-related response. The Big Model can reason about these relationships and identify a corresponding learning objective. In the general framework, such objectives may lead to perception, interaction-selection, action, or outcome-prediction models. The present experimental realization focuses on measurement selection, task prediction, and the applicability of learned task capabilities; more general action and outcome models are possible extensions rather than demonstrated results.

\subsubsection{Discovery and improvement of interaction methods}

A thinking result may include a hypothesis, a proposed interaction method, suggested applicability conditions, and a learning plan. A method specifies how available robot operations are combined to obtain information. For instance, instead of repeatedly using color observations, the Big Model may suggest a close-range observation under side illumination, or touching an object and waiting before reading a material-related signal. These operations may expose properties that are not observable through the initial sensing method. The Big Model is provided with the available physical interfaces, but not with the correct configuration for a given task.

A proposed method is first stored as a candidate rather than accepted as knowledge. The Small Model selects candidates for execution, collects evidence within its interaction budget, and may further modify a candidate by varying its control parameters. Same-object comparison measurements can also be conducted to investigate whether a modified method provides more useful information than the current method. When evidence is insufficient, a promising candidate can be retained for continued investigation instead of being replaced whenever another proposal arrives. Candidate records, including their origins, revisions, and outcomes, are retained for subsequent learning and reasoning.

\subsubsection{Learning which interaction methods to use}

The first form of Small Model growth concerns the acquisition of a better method-selection capability. A lightweight model learns the utility of a measurement method from its observable situation, physical parameters, and available feedback. Consequently, the Small Model need not consult the Big Model whenever a similar measurement decision is required. In the experimental realization, the utility target for method $p$ in situation $s$ is
\begin{equation}
    u(s,p)=A_{p}(s)-A_{0}(s)-\lambda c(p),
    \label{eq:model_utility}
\end{equation}
where $A_{p}$ and $A_{0}$ are the prediction accuracies obtained from the candidate measurement signal and the initial color signal, respectively, on later labeled observations after calibration using earlier observations. The term $c(p)$ is the physical operation cost of the method, and $\lambda$ balances accuracy gain against this cost. A random-forest regressor learns to estimate this utility from method parameters and observable contextual features. The conditions suggested by the Big Model may contribute to these features, but do not constitute verified rules or absolute restrictions. Historical utility, uncertainty about insufficiently tested methods, and the need to consolidate evidence also influence candidate selection. Thus, the Small Model learns from the consequences of executing methods instead of merely imitating the Big Model's choices.

\subsubsection{Learning and validating new task capabilities}

The second form of growth concerns the task model itself. If a candidate provides useful observations, the robot continues to collect suitable learning material. A new task model can incorporate the additional information while retaining the initial inputs. For example, a classifier that initially uses color can be extended to use both color and the signal obtained by a newly discovered optical measurement. The learning algorithm and model architecture may be selected according to the task; the present implementation uses logistic regression for this extension. It does not require the Big Model to generate an arbitrary new neural-network architecture.

The Big Model's role in training is to help determine what capability should be learned, what information the learner requires, and how suitable learning material can be obtained through interaction. The Small Model then uses verified observations and feedback to carry out the numerical training. If an existing model can incorporate the new information without compromising its established capability, it may be extended or retrained. If the new relationship requires a distinct input representation, applicability condition, or function, a separate lightweight model may instead be formed. Thus, the framework permits both growth within an existing model and growth through the formation of additional models. The latter is realized in the experiments by constructing separate task classifiers for relevant locations or object families and measurement configurations.

For a candidate model $M$ and a reference model $M_0$ trained using the original input information, acceptance requires
\begin{equation}
\begin{split}
 A_D(M)&\geq\tau,\\
 A_D(M)-A_D(M_0)&\geq\delta,\qquad
 D\in\{D_v,D_c\},
\end{split}
\label{eq:model_acceptance}
\end{equation}
where $D_v$ and $D_c$ are separate validation and confirmation sets, $\tau$ is the minimum absolute accuracy, and $\delta$ is the required improvement. Training, validation, and confirmation material is obtained from distinct, chronologically ordered interaction episodes with released feedback. Both criteria must hold on both sets. This prevents a large relative improvement over a poor reference from being sufficient on its own. The exact training configuration and thresholds are given in Section~\ref{sec:verification}.

After acceptance, the task model is stored together with its associated measurement method, observable context, and validation history. The Small Model can subsequently execute the measurement and apply the trained model without requesting a new LLM inference. If the candidate does not pass validation, the evidence is retained and may support further measurement, method modification, or Big Model reflection. In this way, thinking can promote both learning \emph{how to obtain information} and learning \emph{how to use that information to perform a task}. Moreover, a successful capability may be the result of a chain of revisions by both components, rather than a direct implementation of the original Big Model proposal.

\subsection{Continual Interaction and Cumulative Model Growth}
\label{sec:model_growth}

Continued environmental interaction is a basic source of growth, rather than merely a response to task failure. While the Big Model is thinking, the robot can continue to observe and interact using its existing small models. New objects, unfamiliar properties, unexpected outcomes, and opportunities to improve successful interactions can provide further subjects for thinking and learning. In this manner, the scope of what the robot can understand and do need not remain limited to the capabilities specified at initialization.

Continued environmental interaction creates two complementary opportunities for growth. First, the relationship between observations and task outcomes may change for an already encountered object or situation. Second, the robot may encounter new objects and phenomena with characteristics not covered by its existing knowledge. The former requires adaptation and possible reuse of historical capabilities; the latter requires an expansion of the robot's knowledge rather than the replacement of all previous capabilities.

For adaptation, the model monitors the actual performance of deployed capabilities using released feedback. A previously effective capability may be suspended when its performance deteriorates. Its parameters, associated interaction method, and historical evidence are preserved. When subsequent observations suggest that the capability may again be useful, the robot can collect fresh evidence and revalidate the archived model. If the revalidation succeeds, the existing parameters can be reactivated rather than learned again from the beginning. The Small Model may also learn an estimate of the current benefit of using a deployed capability, using feedback from past predictions. These mechanisms separate the existence of learned knowledge from its current applicability.

Learning materials likewise need to be selected according to the situation in which they were obtained. Combining observations associated with conflicting environmental relationships may obscure a previously informative property. In changing environments, a detector of changes in the observed relationship between a measurement and its released task feedback can identify a boundary for subsequent evidence selection. A recent evidence interval is then considered for learning, while the older observations remain archived. In environments where different observable object families retain stable properties, family-specific records can instead be used without temporal relation segmentation. Neither mechanism receives a hidden environmental phase label.

For cumulative growth, let $\mathcal{S}_t=\{M_1,\ldots,M_{n_t}\}$ denote the small-model repertoire at time $t$, where $M_j$ is a learned lightweight model and $n_t$ is the number of models retained at that time. Each model is associated with its function, required observations or interaction method, and the evidence supporting its applicability. The repertoire may expand by adding a validated capability, or become more capable by improving an existing member. Suspended models remain in the repertoire but are not necessarily available for routine execution. Growth therefore concerns the range and reliability of the available capabilities, rather than an unconditional increase in model count.

The robot first attempts to handle an encountered object using available knowledge. If the existing capability is insufficient, it can examine previously discovered but unverified methods, collect further measurements, or benefit from new Big Model thinking. The newly validated capability is added to the repertoire without discarding capabilities for earlier objects. Different task models may therefore coexist, and an existing method may be reused to learn a capability for another situation. This permits the robot to increase both the depth of its knowledge of individual objects and the breadth of objects it can handle reliably.

Crucially, the repertoire is not only the outcome of learning but also a resource for subsequent environmental interaction. A robot that has learned when a particular measurement is informative can use its method-selection model to choose that interaction without requesting new LLM reasoning. Its task model can then interpret the resulting observation. These capabilities may be reused when investigating another object or a different property, while their successes, limitations, and newly observed consequences provide further experiences for both components. For example, a previously learned way of obtaining a material-related signal may be considered when the robot encounters an unfamiliar object, even though its predictive relationship with the new task must be verified locally.

This establishes a continual growth cycle: interaction produces new experiences; the Big Model reasons about the experiences and possible learning objectives; the Small Model collects evidence, learns, and validates additional capabilities; and the resulting repertoire supports further interaction. Multiple small models with different functions can contribute to this process according to their applicability, without requiring the Big Model to direct each individual action. As the repertoire expands, it can potentially enable interactions and discoveries that were not possible with the initial capabilities. Such expansion of interaction opportunities is a design objective of the framework, not a guarantee that every newly formed model will lead to more discoveries.

The current verification evaluates the acquisition and independent use of new task models, restoration of archived knowledge, and knowledge accumulation under continued encounters with additional object families. It demonstrates the conversion of interaction experience and Big Model thinking into reusable capabilities and their subsequent use. However, the additional object families are presented by the simulated environment; their discovery is not contingent on the robot's learned abilities. The experiments therefore verify important parts of the growth cycle, but do not yet establish that a growing model repertoire causes the robot to discover more previously inaccessible objects, or that it can autonomously invent unrestricted learning algorithms.

\section{Verification}
\label{sec:verification}

The experiments investigate whether an independently operating large model (Big Model) can contribute to the continual formation and growth of a repertoire of small models (collectively referred to as the Small Model) through interaction with the environment. Two complementary questions are considered: whether the robot can acquire, suspend, and recover task knowledge as environmental conditions change, and whether it can accumulate knowledge as new kinds of objects become available without losing its previous capabilities. The emphasis is on the conversion of thinking results into independently usable task models rather than on the number of hypotheses generated by the Big Model. Model deployments, subsequent use, and the breadth of mastered object families provide complementary evidence of growth; a deployment count is not necessarily the number of distinct simultaneously active capabilities.

\subsection{Experimental Environment and Evaluation Protocol}
\label{sec:verification_setup}

\subsubsection{Simulated robot and sensing environment}
A simulated autonomous inspection robot repeatedly encounters objects and makes a binary task decision for each object instance. The decision may represent, for example, whether an inspected object satisfies a task-related condition. The robot initially has a color-based decision capability but does not know which other physical properties are informative. It can change its observation distance (wide or macro), illumination (normal or side), orientation (unrotated or rotated by $90^\circ$), contact and waiting conditions, and the readout type (color, optical, material, or edge). Combining these controls yields 128 possible measurement configurations. A configuration determines the quality of a simulated binary measurement signal; it does not directly reveal the correct task label. For instance, a side-lit macro observation can reveal an informative optical signal when ordinary color is uninformative. These relationships are implemented by the simulator and are not supplied to the learning components. The experiments concern active sensing and decision learning in a synthetic environment, not image recognition from real camera data or physical robot navigation.

Two scenarios are constructed. In the \emph{changing environment}, the robot alternates between an orchard and a market. Initially, color is informative, with a location-dependent association. Subsequently, optical measurements become informative, then contact-based measurements become informative, and finally the optical relationship returns. Actual change points are perturbed around the nominal phase boundaries and are not announced to either model. The locations remain observable. This scenario tests adaptation to a changing relationship between measurements and decisions, as well as the reuse of previously acquired knowledge.

In the \emph{expanding environment}, the robot first encounters a color-informative object family. Additional families are introduced successively; previous families remain available and their sensing relationships remain unchanged. Object identities are observable upon encounter, but the informative measurement configuration for each new family is hidden. The assignment of optical, contact/material, and edge/orientation relationships to the new families is determined by the environmental seed. At the end, all families are encountered again. This scenario distinguishes the accumulation of additional knowledge from the replacement of knowledge following environmental change. Encounters are generated by a common simulator schedule, rather than by a navigation policy learned by the robot.

\subsubsection{Models and experiment configuration}
The Big Model uses live DeepSeek-v4-flash requests. It receives public observations, the Small Model's released feedback, previous hypotheses and their outcomes, and optionally queries a simulated peer's historical measurement experience. It produces hypotheses about measurement configurations, conditions of applicability, and possible learning plans. These are candidates, not labels or compulsory actions. The Small Model uses a random-forest regressor to estimate measurement utility and a logistic-regression classifier with color and the acquired measurement signal to implement newly learned task knowledge. The measurement-utility forest uses 32 trees, a maximum depth of 6, and a minimum leaf size of 2; the task classifier uses $C=2$ and at most 300 iterations. A separate random-forest applicability estimator (32 trees, maximum depth 4, minimum leaf size 2) can use previous prediction gains to decide whether to use a deployed classifier. The original color-based decision rule is also retained and updated using released feedback. The sensing-policy and task-learning components use actual simulated interaction results rather than labels generated by the LLM.

Each episode contains several inspection cases, and the corresponding task labels become available only after a fixed delay. A sensing configuration is charged for its readout and additional control operations. Optional same-object comparison measurements are charged within the same per-episode budget. The Big Model's reasoning request runs in a separate worker while the Small Model acts on its current knowledge. Proposals are incorporated at the end of a scheduling block; the Small Model does not need a new LLM result before every action, although the next block can wait for a pending result. During designated student-only windows, new LLM reasoning and auxiliary comparison measurements are disabled, but the Small Model may still learn from mature feedback and reuse previously received proposals.

Table~\ref{tab:verification_config} summarizes the shared settings. Candidate task classifiers use chronological, nonoverlapping training, validation, and confirmation episodes. Accepted new models in these runs use the split in the table. Both validation and confirmation must meet the absolute accuracy criterion and improve upon a color-only classifier fitted on the same training material. For the changing-environment method, a public-feedback relation-change detector may separate the recent evidence for a location and measurement configuration; subsequent validation and confirmation must be obtained after detection. The archived experience and previously trained models are retained. The expanding-environment implementation disables this relation-change detector because the object families have stable relationships, while retaining the other learning and evidence-collection mechanisms. Thus, the proposed implementations share the same growth architecture but use different context-handling settings appropriate to the scenarios.

\begin{table}[!t]
\centering
\caption{Experimental configuration.}
\label{tab:verification_config}
\footnotesize
\begin{tabular}{lrr}
\toprule
Setting & Changing & Expanding \\
\midrule
Episodes & 288 & 360 \\
Nominal phase length & 72 & 72 \\
Cases per episode & 6 & 6 \\
Feedback delay (episodes) & 2 & 2 \\
Sensing budget per episode & 36 & 36 \\
Thinking block (episodes) & 4 & 4 \\
Student-only period & 24 & 24 \\
Student-only window & 6 & 6 \\
Recent evidence window & 42 & 42 \\
Fresh evidence window & 26 & 26 \\
Train/validation/confirmation & 6/2/2 & 6/2/2 \\
Minimum accuracy (\%) & 75 & 75 \\
Minimum gain (pp) & 10 & 10 \\
LLM temperature & 0.2 & 0.2 \\
LLM maximum output tokens & 8192 & 8192 \\
Environmental seed & 0 & 0 \\
\bottomrule
\end{tabular}
\end{table}

\subsubsection{Comparison methods and metrics}
The changing-environment experiment compares \emph{Proposed} with \emph{NoContext}, which disables the relation-change-based selection of evidence and corresponding contextual information in reasoning, while retaining the main interaction, learning, and model-reuse mechanisms. In the expanding environment, \emph{Proposed} uses the shared learning architecture described above, \emph{NoBig} replaces LLM proposals with randomly sampled legal measurement configurations while retaining adaptive measurement selection, local revisions, training, validation, and knowledge reuse, and \emph{FixedColor} restricts the robot to its original color measurement but allows its color-based decision rule to adapt using feedback. NoBig is therefore a learning-capable non-LLM comparator, not a purely random action policy. It does not query peers, and its random proposals are available immediately rather than at the end of a reasoning block. The comparisons use common task and encounter streams within each scenario and identical per-episode budget limits, but realized sensing costs and proposal histories need not be equal.

Online accuracy is the proportion of correct predictions among cases whose delayed labels have arrived before the run ends. Student-only accuracy uses the subset generated during the designated windows. A stronger independent evaluation, \emph{frozen accuracy}, fixes the model parameters, measurement configuration, and color fallback at each snapshot, evaluates new case indices after the online run, and never returns the test outcomes to the learner. In the changing environment, the snapshots are taken at the beginning of student-only windows and tested on cases from the corresponding time intervals. In the expanding environment, additional snapshots evaluate each encountered family on 96 independent cases at each phase boundary, including the final boundary. An object family is counted as mastered if its frozen accuracy reaches the minimum accuracy in Table~\ref{tab:verification_config}; the number of mastered families and their equally weighted mean accuracy measure the breadth and quality of knowledge. A newly verified family also requires a deployed new task model and the minimum improvement over its color fallback. Sensing costs of the offline evaluations are recorded separately from online interaction. Finally, a sealed simulator audit identifies whether the executed primary configuration is informative in the actual environment. This indicator is used only for analysis and never for online decisions.

All results below are from a single completed environmental seed for each method. LLM-enabled methods use distinct live reasoning trajectories; the comparisons are descriptive and do not establish statistical significance or isolate every mechanism's causal effect.

\subsection{Adaptation and Reuse of Knowledge in a Changing Environment}
\label{sec:verification_drift}

The first experiment examines whether the Big Model's discoveries can be converted into useful Small Model knowledge and whether the knowledge can be reused when a previously encountered sensing relationship returns. It also assesses the consequence of selecting learning material according to changes in the observed measurement--outcome relationship rather than treating all recent samples from the same location as equally applicable.

\begin{table}[!t]
\centering
\caption{Changing environment: overall performance and cost. Accuracy and informative-sensing rate are percentages.}
\label{tab:verification_drift}
\footnotesize
\begin{tabular}{lrr}
\toprule
Metric & Proposed & NoContext \\
\midrule
Released task cases & 1716 & 1716 \\
Student-only cases & 420 & 420 \\
Frozen test cases & 432 & 432 \\
Online accuracy & 65.73 & 60.37 \\
Student-only accuracy & 69.52 & 62.38 \\
Frozen accuracy & 70.14 & 58.10 \\
Frozen color fallback accuracy & 58.10 & 58.10 \\
Informative primary sensing (\%) & 54.51 & 22.92 \\
New classifier deployments & 3 & 0 \\
Model-use episodes (mature labels) & 47 & 0 \\
Old-model reactivations & 1 & 0 \\
Detected relation changes & 4 & 0 \\
Completed LLM reasoning cycles & 48 & 48 \\
Episodes with pending reasoning & 192 & 192 \\
Peer-history queries & 21 & 21 \\
LLM prompt tokens & 272613 & 283202 \\
LLM completion tokens & 141707 & 159022 \\
Online sensing operations & 6740 & 6382 \\
Offline frozen-test operations & 864 & 432 \\
\bottomrule
\end{tabular}
\end{table}

As shown in Table~\ref{tab:verification_drift}, Proposed achieves 65.73\% online accuracy and 69.52\% student-only accuracy, improving upon NoContext by 5.36 and 7.14 percentage points (pp), respectively. It deploys 3 new classifiers and reactivates 1 archived classifier, whereas NoContext deploys none. Both methods complete 48 LLM reasoning cycles and make 21 peer-history queries. The action log records 192 episodes of Small Model activity while an LLM request remains unfinished in Proposed. This is an implementation-level observation of concurrent activity, not a timing benchmark against a sequential architecture. The actual proposals and sensing activities differ between the methods. Proposed also consumes more online sensing operations. The result therefore supports the feasibility of the learning and recovery mechanism but is not an equal-realized-cost demonstration of the superiority of context separation.

\begin{figure}[!t]
\centering
\includegraphics[width=\columnwidth]{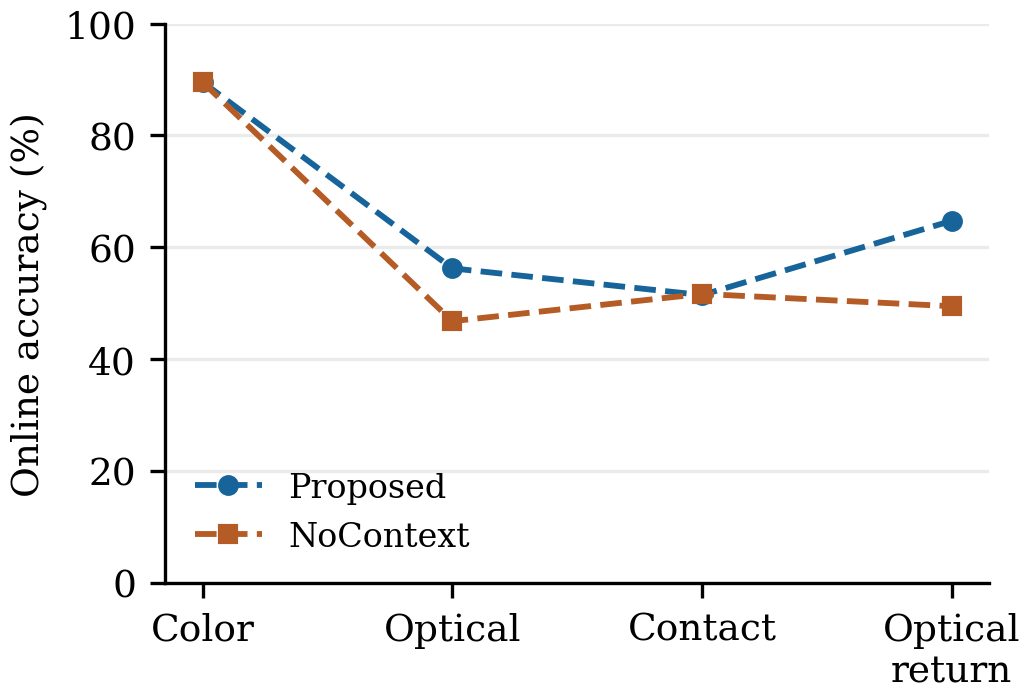}
\caption{Online accuracy by hidden environmental phase. Each point aggregates the released task outcomes in the corresponding phase; dashed segments connect phase summaries rather than episode-level measurements.}
\label{fig:verification_drift}
\end{figure}

\begin{table}[!t]
\centering
\caption{Changing environment: phase-wise task accuracy and informative primary sensing. The sensing counts use all executed episodes, whereas accuracy uses released labels only.}
\label{tab:verification_phases}
\footnotesize
\setlength{\tabcolsep}{4pt}
\begin{tabular}{lrrrrr}
\toprule
& & \multicolumn{2}{c}{Accuracy (\%)} & \multicolumn{2}{c}{Informative} \\
\cmidrule(lr){3-4}\cmidrule(lr){5-6}
Phase & Episodes & Prop. & NoCtx & Prop. & NoCtx \\
\midrule
Color & 78 & 89.53 & 89.53 & 66 & 66 \\
Optical & 67 & 56.22 & 46.77 & 44 & 0 \\
Contact & 81 & 51.44 & 51.65 & 0 & 0 \\
Optical return & 62 & 64.72 & 49.44 & 47 & 0 \\
\bottomrule
\end{tabular}
\end{table}

Fig.~\ref{fig:verification_drift} and Table~\ref{tab:verification_phases} show where the improvements arise. Both methods perform similarly in the initial color-informative phase. When the optical relationship becomes informative, Proposed achieves 56.22\% compared with 46.77\% for NoContext. When this relationship returns later, the corresponding accuracies are 64.72\% and 49.44\%. Neither method executes an informative primary configuration in the intervening contact phase. Thus, the learned optical knowledge improves behavior in the appropriate environment but does not solve every new sensing problem.

\begin{table}[!t]
\centering
\caption{Paired predictions on cases where Proposed used a new classifier in the changing environment. The color fallback and learned classifier are evaluated on identical cases.}
\label{tab:verification_paired}
\footnotesize
\begin{tabular}{lrrr}
\toprule
Phase & Cases & Color (\%) & Learned (\%) \\
\midrule
First optical & 84 & 46.43 & 91.67 \\
Contact & 72 & 47.22 & 45.83 \\
Optical return & 126 & 49.21 & 92.86 \\
\midrule
All frozen covered cases & 144 & 48.61 & 84.72 \\
\bottomrule
\end{tabular}
\end{table}

Table~\ref{tab:verification_paired} compares predictions on the same cases. In the first optical phase, the learned classifier raises accuracy from 46.43\% to 91.67\%; when the optical relationship returns, it raises accuracy from 49.21\% to 92.86\%. Conversely, applying an optical classifier during the contact phase can be detrimental. This demonstrates why knowledge acquisition must be accompanied by an assessment of its current applicability.

The model lifecycle in the market gives a concrete example. As reported in Table~\ref{tab:verification_recall}, an optical classifier is deployed at episode 119 and suspended at episode 167 after its observed performance deteriorates. The archived classifier is subsequently tested on new validation and confirmation material, and is reactivated at episode 255 with 100\% and 100\% accuracy on the two sets. The reactivated model retains the original trained parameters rather than requiring a newly fitted classifier. This is evidence of knowledge reuse, although a separate no-recall comparison would be required to quantify any savings relative to relearning.

\begin{table}[!t]
\centering
\caption{Lifecycle of the recovered market classifier (episode indices).}
\label{tab:verification_recall}
\footnotesize
\begin{tabular}{lrr}
\toprule
Event & Episode(s) & Accuracy (\%) \\
\midrule
Initial deployment & 119 & -- \\
Suspension & 167 & -- \\
Revalidation material & 238--239 & 100 \\
Reconfirmation material & 252--253 & 100 \\
Reactivation & 255 & -- \\
\bottomrule
\end{tabular}
\end{table}

Most importantly, the independent frozen tests include 432 cases in total. Proposed obtains 70.14\% accuracy compared with 58.10\% for NoContext. On the 144 cases covered by frozen new classifiers, the accuracy increases from 48.61\% for the same frozen color fallback to 84.72\%, a gain of 36.11 pp (Table~\ref{tab:verification_paired}). No LLM inference or parameter update takes place in these tests. Consequently, the gain provides direct evidence that the acquired task knowledge resides in the Small Model rather than depending on a contemporaneous answer from the Big Model. The covered subset is conditional on the availability of a deployed model and should not be confused with performance over the entire test set.

\subsection{Cumulative Knowledge Growth through Continued Interaction}
\label{sec:verification_growth}

The second experiment investigates whether encountering more kinds of objects can lead to an expanding portfolio of independently usable task knowledge and associated interaction methods. In particular, it examines whether experience obtained from new sensing operations can be converted into additional small models while earlier capabilities remain available. Unlike the previous scenario, the informative relationships of old object families do not disappear when new families are introduced. A system that merely replaces its current classifier may therefore fail to retain earlier capabilities. The robot is evaluated on all encountered families, including those for which it has not yet learned a new model.

\begin{table}[!t]
\centering
\caption{Expanding environment: overall results and online interaction costs. Frozen accuracy is the final family-balanced accuracy (\%).}
\label{tab:verification_growth}
\footnotesize
\setlength{\tabcolsep}{3.5pt}
\begin{tabular}{lrrr}
\toprule
Metric & Prop. & NoBig & FixedColor \\
\midrule
Released task cases & 2148 & 2148 & 2148 \\
Student-only cases & 528 & 528 & 528 \\
Online accuracy (\%) & 68.06 & 69.37 & 65.32 \\
Student-only accuracy (\%) & 66.10 & 65.15 & 63.26 \\
Final frozen accuracy (\%) & 69.01 & 79.43 & 57.81 \\
Encountered families & 4 & 4 & 4 \\
Mastered families & 2 & 3 & 1 \\
Newly verified families & 1 & 2 & 0 \\
New classifier deployments & 1 & 2 & 0 \\
Model-use episodes (mature) & 27 & 37 & 0 \\
Online sensing operations & 7976 & 10144 & 2160 \\
Family-test operations & 1920 & 2592 & 1344 \\
Periodic frozen-test ops. & 756 & 900 & 540 \\
LLM reasoning cycles & 60 & 0 & 0 \\
Episodes with pending reasoning & 240 & 0 & 0 \\
Peer-history queries & 10 & 0 & 0 \\
LLM prompt tokens & 354853 & 0 & 0 \\
LLM completion tokens & 152419 & 0 & 0 \\
\bottomrule
\end{tabular}
\end{table}

\begin{figure}[!t]
\centering
\includegraphics[width=\columnwidth]{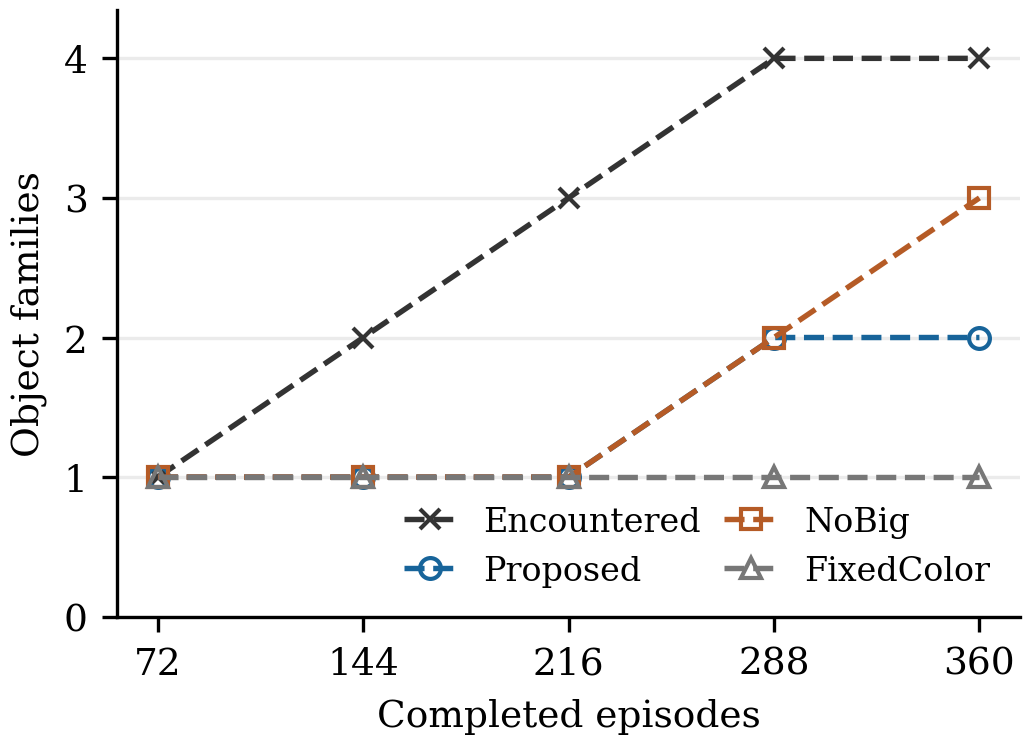}
\caption{Growth in the number of mastered object families evaluated on independent frozen tests. The encountered-family count is identical for all methods. Dashed lines connect evaluation snapshots.}
\label{fig:verification_growth}
\end{figure}

As shown in Fig.~\ref{fig:verification_growth}, the number of encountered families increases from 1 at the first evaluation to 4 at the final evaluation. FixedColor remains able to master only 1 family; Proposed expands its mastered set to 2, while NoBig reaches 3. The final frozen accuracies in Table~\ref{tab:verification_growth} are 57.81\%, 69.01\%, and 79.43\% for FixedColor, Proposed, and NoBig, respectively. Thus, continued measurement discovery and learning permit additional knowledge to be acquired beyond the original sensing capability. The non-LLM method also demonstrates cumulative growth and attains the highest final coverage in this run.

\begin{table}[!t]
\centering
\caption{Final frozen accuracy (\%) and informative primary sensing episodes by object family. Each family is tested on 96 held-out cases. The family index is an observed identity, not its hidden sensing relationship.}
\label{tab:verification_objects}
\footnotesize
\setlength{\tabcolsep}{3.5pt}
\begin{tabular}{lrrrrr}
\toprule
& \multicolumn{3}{c}{Frozen accuracy (\%)} & \multicolumn{2}{c}{Informative} \\
\cmidrule(lr){2-4}\cmidrule(lr){5-6}
Family & Prop. & NoBig & Fixed & Prop. & NoBig \\
\midrule
A (color) & 88.54 & 88.54 & 88.54 & 110/126 & 106/126 \\
B (optical) & 52.08 & 93.75 & 52.08 & 0/90 & 19/90 \\
C (edge) & 42.71 & 42.71 & 42.71 & 0/72 & 0/72 \\
D (material) & 92.71 & 92.71 & 47.92 & 40/72 & 69/72 \\
\bottomrule
\end{tabular}
\end{table}

Table~\ref{tab:verification_objects} identifies the acquired abilities. Proposed retains 88.54\% accuracy on family A, identical to the original color-based comparator on the same final cases. For family D, it improves accuracy from 47.92\% to 92.71\%, a gain of 44.79 pp. This is a newly acquired and independently verified capability, while the previously mastered family remains mastered. Neither learning method masters family C. The comparison therefore measures the partial accumulation of knowledge, not complete mastery of the environment.

\begin{table}[!t]
\centering
\caption{Trace of measurement discovery and learning in Proposed for family D.}
\label{tab:verification_lineage}
\footnotesize
\begin{tabular}{lp{0.60\columnwidth}}
\toprule
Episode & Event \\
\midrule
8 & Big Model proposes macro, side light, contact, wait, and material readout. \\
154 & Feedback-guided Small Model changes side light to normal light, retaining the other controls. \\
248 & Task classifier is deployed; validation 91.67\%, confirmation 100\%. \\
360 & Final frozen evaluation: 92.71\% accuracy. \\
\bottomrule
\end{tabular}
\end{table}

The lineage in Table~\ref{tab:verification_lineage} illustrates the cooperation between the models. The Big Model proposes a contact-and-wait material measurement at episode 8. At episode 154, a feedback-guided local revision by the Small Model changes the illumination condition. This derivative configuration eventually yields a deployed model at episode 248, with 91.67\% validation accuracy and 100\% confirmation accuracy. This case is not merely direct imitation of an LLM answer: the final sensing configuration is an actual modification of an earlier thinking result, and its predictive knowledge is obtained from labeled interaction experience.

\begin{table}[!t]
\centering
\caption{An unsuccessful conversion of an LLM optical proposal into learned knowledge for family B.}
\label{tab:verification_bottleneck}
\footnotesize
\begin{tabular}{lr}
\toprule
Recorded event or quantity & Value \\
\midrule
Initial proposal episode & 0 \\
Auxiliary measurement episode & 89 \\
Auxiliary cases with mature feedback & 6 \\
Signal--label agreements & 5 \\
Pair expiration episode & 126 \\
Primary executions on family B & 0 \\
\bottomrule
\end{tabular}
\end{table}

The limitations of the mechanism are also informative. In Table~\ref{tab:verification_bottleneck}, a macro side-lit optical configuration is proposed at episode 0. It is used on family B only as an auxiliary measurement at episode 89, where 5 of 6 measured signals agree with the released labels. The comparison activity expires at episode 126 without the configuration ever becoming a primary measurement on that family. The candidate is objectively informative according to the sealed environment definition, yet its available evidence is insufficient for model training and independent confirmation. This separates failure to discover a useful hypothesis from failure to allocate adequate interaction opportunities to it. It also explains why more LLM proposals alone need not yield more acquired knowledge.

Proposed uses 7976 online sensing operations, 21.37\% fewer than NoBig's 10144 operations (Table~\ref{tab:verification_growth}), and its student-only accuracy is 0.95 pp higher. However, NoBig acquires more verified families and has higher final frozen accuracy. Its exploration is viable in the finite, structured measurement space; moreover, it incurs no LLM computation. Consequently, the present results do not establish a general efficiency or performance advantage of LLM reasoning over adaptive non-LLM exploration. They demonstrate that an LLM can produce useful measurement hypotheses that the Small Model subsequently revises and internalizes, while identifying candidate scheduling as an important bottleneck.

\subsection{Discussion}

The experiments demonstrate complementary aspects of continual growth: the acquisition and later recovery of a task capability under environmental change, and the accumulation of additional capabilities while retaining earlier knowledge as objects are added. The actual use of deployed classifiers shows that the acquired models participate in subsequent sensing and task decisions, rather than merely being stored. These results support the formation and reuse of a model repertoire on a limited scale; they do not demonstrate the emergence of a large collection of heterogeneous action models or establish the entire feedback loop in which improved interaction capabilities themselves create additional discovery opportunities. The frozen tests distinguish previously learned knowledge from immediate LLM assistance. The simulations also show a limitation of the cooperation mechanism: useful proposed measurements do not necessarily receive sufficient trials. The evidence supports the feasibility of the framework, not superiority over every non-LLM search strategy. Further multi-seed trials, shared-proposal ablations, a no-growth comparator, and a no-recall comparator are needed to assess robustness, isolate mechanism-specific contributions, and quantify savings from knowledge reuse. Finally, because object encounters are externally scheduled and the experiments use logical blocks of concurrent reasoning and action, they do not establish autonomous discovery of new physical objects, unrestricted parallel execution, or real-world robotic performance.

\section{Conclusion} \label{sec_conclusion}
This paper proposed an independent and parallel large--small model collaboration framework for the continual growth of autonomous robots. Through continuous environmental interaction, the large model autonomously thinks about new objects, interaction phenomena, and accumulated experiences, promoting the formation and improvement of small models. These models enhance subsequent environmental interaction, enabling further knowledge acquisition and capability growth. Experimental results demonstrated the feasibility of transforming reasoning outcomes into independently usable small models, accumulating new knowledge while retaining existing capabilities, and reactivating previously learned models under changing environmental conditions. Future work will focus on improving the efficiency of thinking-driven learning and validating the continual growth mechanism in more complex and open-ended physical environments.


\ifCLASSOPTIONcaptionsoff
  \newpage
\fi

\bibliographystyle{IEEEtran}
\bibliography{ref}

%

\begin{IEEEbiography}{Hong Su}
  received the MS and PhD degrees, in 2006 and 2022, respectively, from Sichuan University, Chengdu, China. He is currently a researcher of Chengdu University of Information Technology Chengdu, China. His research interests include LLM and autonomous robots.
\end{IEEEbiography}




\end{document}